# 语义万维网中基于符号变换的超协调表演算[*]

张小旺[+], 肖国辉, 林作铨

(北京大学 数学科学学院 信息科学系, 北京 100871)

# A Paraconsistent Tableau Algorithm Based on Sign Transformation in Semantic Web[*]

ZHANG Xiao-Wang[+], XIAO Guo-Hui, LIN Zuo-Quan

(Department of Information Science, School of Mathematical Sciences, Peking University, Beijing 100871, China)

+ Corresponding author: Phn: +86-10-62757175, E-mail: zxw@is.pku.edu.cn

**Abstract**: In an open, constantly changing and collaborative environment like the forthcoming Semantic Web, it is reasonable to expect that knowledge sources will contain noise and inaccuracies. It is well known, as the logical foundation of the Semantic Web, description logic is lack of the ability of tolerating inconsistent or incomplete data. Recently, the ability of paraconsistent approaches in Semantic Web is weaker in this paper, we present a tableau algorithm based on sign transformation in Semantic Web which holds the stronger ability of reasoning. We prove that the tableau algorithm is decidable which hold the same function of classical tableau algorithm for consistent knowledge bases.

**摘 要**: 语义万维网作为一个开放、不断更新而且相互协作的环境,经常会包含一些不协调的或不精确的信息。众所周知,描述逻辑是语义万维网重要的逻辑基础,然而描述逻辑缺乏处理不协调或不完全信息的能力,致使语义万维网缺乏足够的超协调推理能力。在本文中,我们提出一种基于符号变换的具有强推理能力的超协调表演算。证明了该算法是可判定的,而且在处理协调的本体情况下该算法的结果与经典逻辑系统的表演算是一致的。

**关键词**: 语义万维网;不协调处理;超协调推理;表演算

中图法分类号: TP301　　　　文献标识码: A

## 1 引言

语义万维网（Semantic Web）作为万维网的进一步发展由蒂姆·伯纳斯-李(Tim Berners-Lee)在 1998 年提出[1]。与万维网存储大量数据相比,语义万维网技术注重的是挖掘这些数据背后的语义信息,而不仅仅只能以数据输入的形式检索信息。当前,语义万维网技术正在迅速发展[2,3,4],各种语义万维网技术的应用得到了广泛开发[5]。W3C 已经提出多种语义万维网知识表示的标准语言,如 RDF(S), OWL。这些语言之所以成为标准语义万维网语言,是因为它们具有严格的逻辑基础——描述逻辑（Description Logic）。因为语义万维网是一个开放、不断更新而且相互协作的环境,所以要求语义万维网上面的数据都是协调完美的是不现实的[14]。

然而描述逻辑继承了一阶逻辑的平凡性，即从逻辑角度讲，描述逻辑是无法处理矛盾信息。因为，研究可以容忍不协调本体的方法，对于语义万维网技术在多数据源上的实用性具有重要意义。

不协调处理一直以来是人工智能领域中的研究问题，相应的成果近来被应用到描述逻辑系统上，基本包含了两种不同的方法：第一类方法是基于矛盾信息显示了系统建模错误的观点，从而需要修复来维持一个协调的知识库。这种处理逻辑不协调信息的方法也包括许多工作[6,7]，然而这种方法的缺点是因为删除或调整而造成有用信息的丢失。第二类办法是通过定义非经典的推理关系，使得可以非平凡地对不协调知识进行推理，这种方法称为超协调处理。这种办法是出于"矛盾的世界"的观点，即在真实数据上出现矛盾是一种自然合理的现象。超协调逻辑使用的是不同于经典二值逻辑的非经典逻辑蕴涵关系。超协调逻辑的基本目标是限制矛盾部分对其他信息的污染，从而制止了经典逻辑的矛盾膨胀性[15,16,17]。目前已有许多这个方面工作[8,9,15]，然而超协调逻辑推理能力比较弱，经典逻辑中很多重要的推理规则都不满足。例如，Yue 提出的四值描述逻辑[8,15]不满足三条基本推理规则：分离规则(MP)，拒取推理(MT)和析取三段论(DS)。针对四值描述逻辑的不足，Zhang 等提出准经典(QC)描述逻辑[10,11]满足了这三条基本推理规则从而增加了推理能力。然而准经典描述逻辑却丢失其它推理优点，如不满足排中律。

因为这些处理不协调方法的不足，自然地，我们期望有一种超协调方法，即能处理不协调又能尽可能保持经典推理系统的能力。然而，迄今仍然是一个开放性问题。为了尝试解决这个问题，我们首先要分析描述逻辑中不协调的信息是如何导致平凡推理。例如：一个本体 $O$=（∅,{$Bird(Tweety)$, ¬ $Bird \sqcup Fly(Tweety)$, ¬ $Fly(Tweety)$}），对于任意查询 $Haswing(Tweety)$调用经典表演算判断 $O \cup \{¬ Haswing(Tweety)\}$的可满足性用来判断是否是 $O \models Haswing(Tweety)$。利用表演算中扩展规则中⊔-规则，可得到两个集合{ $Bird(Tweety)$, ¬ $Bird(Tweety)$, ¬ $Fly(Tweety)$, ¬ $Haswing(Tweety)$}和{ $Bird(Tweety)$, $Fly(Tweety)$, ¬ $Fly(Tweety)$, ¬ $Haswing(Tweety)$}，容易看出这两个集合都是封闭的。所以 $O \cup \{¬Haswing(Tweety)\}$是不可满足的，即 $O \models Haswing(Tweety)$。所以经典表演算面对不协调的本体时，产生平凡推理。我们容易看出，产生平凡推理的原因是由表扩展规则产生了封闭的集合。而造成封闭的原因是因为本体自身是不协调的。于是，在面对不协调的本体时，通过封闭性来决定不满足性显然有缺陷。根据处理不协调两种基本的方法的共同点是避免平凡推理，而对于不协调的信息，要么直接删除要么以矛盾形式存在。从这个观点来看，当本体是不协调的时，研究的主要问题是推理系统还可以继续对那些与不协调的信息无关知识进行推理和判断。在上面例子中，利用⊔-规则得到两个集合中，没有任何信息与¬ $Haswing(Tweety)$有矛盾，即，¬ $Haswing(Tweety)$不是导致这两个集合是封闭的直接原因。容易看出，经典表演算在定义封闭性过于笼统，没有区分前提和结论产生封闭性的差异。

本文工作的出发点是：启发于林作铨和李未提出悖论逻辑的表演算[17]，在描述逻辑中，用标记符号来定义三种冲突以此来讨论产生封闭性的差异，并针对现有的超协调表演算的推理能力的不足而提出一种具有较强的推理能力的超协调表演算。我们继承了悖论逻辑的表演算[17]的优点，使我们的超协调表演算建立在经典描述逻辑表演算推理系统中增加标记来实现推理任务。同时我们重建了悖论逻辑的表演算[17]的符号标记，使得我们的标记更适合描述逻辑的表演算。另外，与悖论逻辑的表演算[15,17]相比，我们讨论了三种冲突与其他现有处理不协调的表演算之间的关联；与基于四值语义的超协调方法[8,15]相比，有着较强的推理能力；与基于准经典语义的超协调方法[10,11]相比，保持了经典逻辑的推理优点，比如满足重言式。

本文的结构如下：在第二节中，简单介绍语义万维网的逻辑基础——描述逻辑的语义和语法；在第三节中，用符号"0"、"1"分别来标记本体知识的假设和查询。定义基于符号变换的表演算的扩展规则和基于符号变换的封闭性，并提出基于符号变换的表演算（称为超协调表演算）；在第四节中，讨论了超协调表演算的性质，证明了该算法的可判定性并重点讨论了在协调情况下与经典推理系统的一致性，最后分析了该算法的复杂度；在第五节中，我们将本文提出的超协调表演算与其它描述逻辑超协调方法进行比较；在最后一节，我们总结文章的工作以及提出未来后续的工作。

## 2 语义万维网的逻辑基础

描述逻辑 *ALC* 是描述逻辑家族的基本成员之一，是我们本文研究的出发点。本节我们简要介绍描述逻辑

A*ALC* 的术语，更详尽内容见[12]。给定一个描述逻辑语言 $\mathcal{L}$：

$$\mathcal{L}=\{\mathcal{N}_C, \mathcal{N}_R, \mathcal{N}_I | \mathcal{N}_C \text{ 是概念名集合}, \mathcal{N}_R \text{ 是角色名集合}, \mathcal{N}_I \text{ 是个体集合}\}.$$

*ALC* 中允许的复杂概念由 $\mathcal{L}$ 通过如下递归定义：（1）全概念⊤，空概念⊥和每个原子概念都是概念；（2）如果 $C,D$ 是概念，那么 $C \sqcup D$, $C \sqcap D$, 和 $\neg C$ 也是概念；（3）如果 $C$ 是一个概念，$R$ 是一个角色，那么 $\forall R.C$ 和 $\exists R.C$ 都是概念。

描述逻辑 *ALC* 的形式化语义是通过解释来定义的。映射 $\cdot^I$ 将每个概念映射为定义域中的一个子集，将每个角色解释为定义域上的一个二元关系。形式化的，一个解释记做 $I=(\Delta^I, \cdot^I)$，包含一个非空定义域 $\Delta^I$ 和一个满足表 1 所示条件的映射 $\cdot^I$。若一个解释 $I$ 使得 $C^I \neq \emptyset$，则 $I$ 称为一个概念 $C$ 的模型。若一个概念 $C$ 存在一个模型，则称为 $C$ 是可满足的。

表 1 描述逻辑 *ALC* 的语法语义

| 构 子 | 语 法 | 语 义 |
|---|---|---|
| 原子概念 $A$ | $A$ | $A^I \subseteq \Delta^I$ |
| 原子角色 $R_A$ | $R$ | $R^I \subseteq \Delta^I \times \Delta^I$ |
| 个体 $I$ | $o$ | $o^I \in \Delta^I$ |
| 全概念 | ⊤ | $\Delta^I$ |
| 空概念 | ⊥ | $\emptyset$ |
| 交 | $C \sqcap D$ | $C^I \cap D^I$ |
| 并 | $C \sqcup D$ | $C^I \cup D^I$ |
| 否定 | $\neg C$ | $\Delta^I \setminus C^I$ |
| 存在量词 | $\exists R.C$ | {x\|存在 y,使得（x,y）$\in R^I$，且 $y \in C^I$} |
| 任意量词 | $\forall R.C$ | {x\|对任意 y,若（x,y）$\in R^I$，则 $y \in C^I$} |

| 公理名称 | 语 法 | 语 义 |
|---|---|---|
| 概念包含 | $C \sqsubseteq D$ | $C^I \subseteq D^I$ |
| 个体概念断言 | $C(a)$ | $a^I \in C^I$ |
| 个体角色断言 | $R(a,b)$ | $(a^I, b^I) \in R^I$ |

一个描述逻辑 *ALC* 的本体 $O$ 包含了成为 ABox 的一组关于个体断言的集合和称为 TBox 的一组概念包含公理的集合，记为 $O$=(TBox,ABox)。表 1 给出了 TBox 和 ABox 公理的具体语义解释。若一个解释 $I$ 满足一个 ABox $\mathcal{A}$ 中的所有断言，则称为 $I$ 是 $\mathcal{A}$ 的模型。若一个 ABox $\mathcal{A}$ 存在一个模型，则称为 $\mathcal{A}$ 是协调的。若一个解释 $I$ 满足一个 TBox $\mathcal{T}$ 中的所有断言，则称为 $I$ 是 $\mathcal{T}$ 的模型。若每个 TBox $\mathcal{T}$ 模型都是概念 $C$ 的模型，则称为 $C$ 是关于 $\mathcal{T}$ 可满足的。若每个 TBox $\mathcal{T}$ 模型都满足 $C \sqsubseteq D$，则称为 $C$ 关于 $\mathcal{T}$ 包含于 $D$，记 $C \sqsubseteq_\mathcal{T} D$。若每个 TBox $\mathcal{T}$ 模型都是 ABox $\mathcal{A}$ 的模型，则称为 $\mathcal{A}$ 是关于 $\mathcal{T}$ 协调的。

**引理 2.1**[12]. 给定一个 ABox $\mathcal{A}$，两个概念 $C,D$ 和一个个体 $a$,有
— $C$ 是可满足的当且仅当存在一个个体 $a \in N_I$ 使得 ABox $\{C(a)\}$ 是协调的；
— $a$ 是 $C$ 关于 $\mathcal{A}$ 的实例当且仅当 $\mathcal{A} \cup \{\neg C(a)\}$ 是不协调的；
— $C$ 包含于 $D$ 当且仅当 $\mathcal{A} \cup \{C \sqcap \neg D(\iota)\}$ 是不协调的，这里 $\iota$ 是一个没有出现在 $\mathcal{A}$ 中的新个体。

**引理 2.2**[13]. 设一个本体 $O=(\mathcal{T}, \mathcal{A})$ 和两个概念 $C,D$。定义：$C_\mathcal{T} := \sqcap_{C_i \sqsubseteq D_i \in \mathcal{T}} \neg C_i \sqcup D_i$。
— $C$ 是关于 $\mathcal{T}$ 可满足的当且仅当 $C \sqcap C_\mathcal{T}$ 是可满足的；
— $C$ 关于 $\mathcal{T}$ 包含于 $D$ 当且仅当 $C \sqcap \neg D \sqcap C_\mathcal{T}$ 是不可满足的；
— $\mathcal{A}$ 是关于 $\mathcal{T}$ 协调的当且仅当 $\mathcal{A} \cup \{C_\mathcal{T}(a) | a \in \cup_\mathcal{A}\}$ 是协调的；

这里$\cup_\mathcal{A}$ 表示所有出现在 $\mathcal{A}$ 中个体组成的集合。

## 3 基于符号变换的超协调表演算

### 3.1 符号变换

在逻辑系统中，因为推理基于一个公式集 $\mathcal{K}$ 和一个公式 $S$ 的二元关系（称为蕴含关系），所以推理存在于两个公式集：一个称为前提，另一个称为结论。对于一个公式，它要么是前提要么不是前提，也就是说，任意的公式关于给定的作为前提的公式集有两个可能性，在本文，我们用两个标记 *0* 和 *1* 来表示这两种可能。

**定义** 3.1. 给定一个公式集 $\mathcal{K}$，设一个公式集 $S$，一个函数 $\varphi_{\mathcal{K}}: S \to \{0, 1\}$ 称为 $S$ 关于 $\mathcal{K}$ 的特征函数，若 $\varphi_{\mathcal{K}}$ 满足：(1) 如果 $\in \mathcal{K}$，$\varphi_{\mathcal{K}}(\ )=1$；(2) 如果 $\notin \mathcal{K}$，$\varphi_{\mathcal{K}}(\ )=0$。此时，$\varphi_{\mathcal{K}}(\ )$ 称为 的标记。

注：1. 特征函数本质上是一个分类函数，即将所有的信息分成两类。

2. 特征函数用 0，1 符号来刻画一个公式与一个公式集合之间的所属关系，所以特征函数也可以称为标记函数，即给每一个公式关于一个公式集合的标记符号。

3. 同一个公式在不同的特征函数中给出的值或符号不一定相同。

在下面的定义中，对于任意给定一个公式集 $\mathcal{K}$，对于任意的公式 $C(a)$ 和它的标记 s，我们可以用一个三元组 $(C,a,s)$ 来表示，这里 $C$ 是一个概念且 $a$ 是一个个体。我们用 $S((C,a,s))$ 表示 $C(a)$ 的标记 s，即 $S((C,a,s))=s$。

在描述逻辑 **ALC** 中，因为不存在角色的否定($\neg R$)构子（这里 $R$ 是一个角色），所以产生的冲突只可能是在概念断言之间。在本文，我们只研究个体概念断言上的符号变换，而个体角色断言由特征函数来定义它的标记。

**定义** 3.2. 给定一个公式集 $\mathcal{K}$，设一个仅含有个体概念断言的公式集 $S$，一个从 $S$ 到 $\mathcal{N}_C \times \mathcal{N}_I \times \{0, 1\}$ 变换 $\mathcal{M}_\mathcal{K}$ 称为 $\mathcal{K}$ 的标记，如果 $\mathcal{M}_\mathcal{K}$ 满足：

(1) $S(\mathcal{M}_\mathcal{K} C(a)) = \varphi_\mathcal{K}(C(a))$；

(2) $S(\mathcal{M}_\mathcal{K} C(a))=s$ 且 $S(\mathcal{M}_\mathcal{K} D(a))=s$，如果 $S(\mathcal{M}_\mathcal{K} C \sqcup D(a)) = s$；

(3) $S(\mathcal{M}_\mathcal{K} C(a))=s$ 且 $S(\mathcal{M}_\mathcal{K} D(a))=s$，如果 $S(\mathcal{M}_\mathcal{K} C \sqcap D(a)) =s$；

(4) $S(\mathcal{M}_\mathcal{K} C(b))=s$ ($b \notin \cup_\mathcal{K}$) 且 $\varphi_\mathcal{K}(R(a,b))=1$，如果对于任意 $x \in \cup_\mathcal{K}$ 有 $\varphi_\mathcal{K}(R(a,x))=0$ 且 $S(\mathcal{M}_\mathcal{K} \exists R.C(a))=s$；

(5) $S(\mathcal{M}_\mathcal{K} C(b))=s$，如果 $\varphi_\mathcal{K}(R(a,b))=1$ 且 $S(\mathcal{M}_\mathcal{K} \forall R.C(a))=s$；

(6) $S(\mathcal{M}_\mathcal{K} C(a))=s$，如果 $S(\mathcal{M}_\mathcal{K} \neg \neg C(a))=s$；

(7) $S(\mathcal{M}_\mathcal{K} C \sqcap D(a))=s$，如果 $S(\mathcal{M}_\mathcal{K} D \sqcap C(a))=s$；

(8) $S(\mathcal{M}_\mathcal{K} C \sqcup D(a))=s$，如果 $S(\mathcal{M}_\mathcal{K} D \sqcup C(a))=s$；

(9) $S(\mathcal{M}_\mathcal{K} \neg C \sqcap \neg D(a))=s$，如果 $S(\mathcal{M}_\mathcal{K} \neg (C \sqcup D)(a))=s$；

(10) $S(\mathcal{M}_\mathcal{K} \neg C \sqcup \neg D(a))=s$，如果 $S(\mathcal{M}_\mathcal{K} \neg (C \sqcap D)(a))=s$；

(11) $S(\mathcal{M}_\mathcal{K} (C \sqcap D) \sqcup (C \sqcap E)(a))=s$，如果 $S(\mathcal{M}_\mathcal{K} C \sqcap (D \sqcup E)(a))=s$；

(12) $S(\mathcal{M}_\mathcal{K} (C \sqcup D) \sqcap (C \sqcup E)(a))=s$，如果 $S(\mathcal{M}_\mathcal{K} C \sqcup (D \sqcap E)(a))=s$；

(13) $S(\mathcal{M}_\mathcal{K} \forall R.\neg C(a))=s$，如果 $S(\mathcal{M}_\mathcal{K} \neg (\exists R.C)(a))=s$；

(14) $S(\mathcal{M}_\mathcal{K} \exists R.\neg C(a))=s$，如果 $S(\mathcal{M}_\mathcal{K} \neg (\forall R.C)(a))=s$；

这里 $s \in \{0,1\}$，$C,D$ 是概念，$R$ 是角色和 $a$ 是个体。

在定义 3.2 中，引入标记目的是为了保持公式在变换过程中标记符号的一致性。(1)表示利用个体概念断言关于给定的公式集的特征值作为标记; (2)—(5)保持公式在本节接下来提出的超协调表演算中标记符号的一致性; (6)—(14)保持公式在转换成否定范式（NNF）后标记符号的一致性。

**定理** 3.2. 若 $C_{NNF}$ 是 $C$ 的 NNF 范式形式，则 $\mathcal{M}_\mathcal{K} C(a) = \mathcal{M}_\mathcal{K} C_{NNF}(a)$。

**证明**：由定义 3.2 直接得出。 □

## 3.2 超协调表演算

在定义基于符号变换的超协调表演算，我们需要引入一些基本概念。定于如下：

**定义** 3.3.

（1）我们用 $\sim C$ 记负概念 $\neg C$ 的 NNF 范式；用 clos($C$) 记包含 $C$ 且在包含和 $\sim$ 变换下是封闭的最小集合；用 clos($\mathcal{A}$):= $\bigcup_{C(a)\in\mathcal{A}}$ clos($C$)，即，clos($\mathcal{A}$) 是所有出现在 $\mathcal{A}$ 中的概念的最小演绎闭包的并。

（2）一个公式集 $\mathcal{K}$ 的森林 $\mathcal{F}_{\mathcal{A}}$ 定义为由根不同的结点和边构成的树的集合。而且每个结点 x 表示一个集合 L(x)⊆clos($\mathcal{K}$)，每条边<x,y>表示一个集合 L(<x,y>)⊆$R_{\mathcal{K}}$，这里 $R_{\mathcal{K}}$ 表示所有出现在 $\mathcal{A}$ 中角色组成的集合。

（3）结点 y 是结点 x 的 $R$-后继如果结点 x 和结点 y 通过边<x,y>连接，即 $R\in$L(<x,y>)。结点 y 是结点 x 的 $R$-前驱如果结点 x 和结点 y 通过边<y,x>连接，即 $R\in$L(<y,x>)。若结点 y 是结点 x 的 $R$-前驱或 $R$-后继，则结点 y 是结点 x 的 $R$-邻居。

（4）结点 x 被 y 直接阻塞当且仅当它的祖先结点都没有被阻塞，而且它有祖先结点 x',y 和 y',使得
— y 不是根节点
— 结点 x 是结点 x'的后继且结点 y 是结点 y'的后继
— L(x)=L(y)且 L(x',)=L(y',)
— L(<x',x>)=L(<y',y>)

结点 x 被间接阻塞当且仅当它的一个祖先结点被阻塞。

（5）给定一个公式集 $\mathcal{K}$ 和一个结点 x，如果存在一个原子概念 $A\in\because C$，使得
— {$A,\neg A$}⊆L(x)且 S($\mathcal{M}_{\mathcal{K}}A$(x))+ S($\mathcal{M}_{\mathcal{K}}\neg A$(x))=2，则称 L(x) 包含一个实冲突；
— {$A,\neg A$}⊆L(x)且 S($\mathcal{M}_{\mathcal{K}}A$(x))+ S($\mathcal{M}_{\mathcal{K}}\neg A$(x))=1，则称 L(x) 包含一个强冲突；
— {$A,\neg A$}⊆L(x)且 S($\mathcal{M}_{\mathcal{K}}A$(x))+ S($\mathcal{M}_{\mathcal{K}}\neg A$(x))=0，则称 L(x) 包含一个内冲突。

我们定义实冲突、强冲突和内冲突都是冲突并规定实冲突为第一类冲突，强冲突和内冲突为第二类冲突。

（6）一个树是完全的，如果该树中任意结点都不能再使用表 2 中的扩展规则。一个树是无冲突的，如果该树中任意结点都不包含冲突。一个树是第一类封闭的(或第二类封闭的)，如果该树中的存在一个结点包含第一类冲突而且任何结点不包含第二类冲突(或第二类冲突)。一个树是第二类封闭的，如果该树中的存在一个结点包含第二类冲突。一个树是封闭的，如果该树是第一类封闭的或第二类封闭的。

（7）一个森林是完全的，如果该森林中任意树都是完全的。一个森林是无冲突的，如果该森林中任意树都是无冲突的。一个森林是封闭的，如果该森林中任意的树都是封闭的而且存在一个树是第二类封闭的。

表 2  描述逻辑 *ALC* 超协调表演算的变换规则

| 规则名称 | 规则描述 |
| --- | --- |
| ⊓-规则 | 条件：(1)$C\sqcap D\in$L(x)，x 不被间接阻塞；(2) {$C,D$}⊈L(x)。<br>动作：L(x):=L(x)∪{$C,D$} |
| ⊔-规则 | 条件：(1)$C\sqcup D\in$L(x)，x 不被间接阻塞；(2) {$C,D$}∩L(x)=∅。<br>动作：$L_1$(x):=L(x)∪{$C$}，$L_2$(x):=L(x)∪{$C$} |
| ∃-规则 | 条件：(1) ∃$R.C\in$L(x)，x 不被阻塞；(2) x 没有 $R$-邻居 y 使得 $C\in$L(y)。<br>动作：生成一个新结点 y，使得 L(<x,y>):={$R$}且 L(y):={$C$}。 |
| ∀-规则 | 条件：(1) ∀$R.C\in$L(x)，x 不被间接阻塞；(2) x 存在 $R$-邻居 y 使得 $C\notin$L(y)。<br>动作：L(y):= L(y)∪{$C$}。 |

接下来，我们基于表 2 中的变换规则提出描述逻辑的超协调表演算。

**算法** 3.1. 描述逻辑 *ALC* 超协调表演算

```
function Paraconsistent-Tableau(...)    return true or false        // 给定一个公式集 𝒦，返回布尔值
    Mitializing
        L(x^i)      ←    {C|C(a^i)∈𝒦}                              // 构造分支的根结点
        L(<x^i,y^i>) ← {R|R(a^i,b^i)∈𝒦}                            // 建立结点之间的边
        ⌐𝒦         ←    (L(x^i), L(<x^i,y^i>))                     // 通过结点和边构造初始森林
        ℳ_𝒦(⌐𝒦)   ←    ℳ_𝒦C(x^i)                                 // 对⌐𝒦中每个结点包含的概念进行标记
    repeat
        ⌐𝒦 ←ApplyingRules(⌐𝒦)                                      // 对⌐𝒦使用表2中的⊓, ⊔, ∃, ∀-规则
        Updating(ℳ_𝒦(⌐𝒦))                                          // 对新的⌐𝒦中每个结点包含的概念进行标记
    until (isComplete(⌐𝒦) )                                         // 判断⌐𝒦是否完全
    if isClosed(⌐𝒦)                                                 // ⌐𝒦是封闭的
        return    true                                              // 返回 ture
    else                                                            // ⌐𝒦 不是封闭的
        return    false                                             // 返回 false
```

在下面定义中，将通过算法 3.1 提出的超协调表演算来刻画一个公式集与一个公式之间的关系。

**定义** 3.4. 给定一个公式集 $\mathcal{K}$ 和一个公式 ，$\mathcal{K}$ 形式推导出 ，记为 $\mathcal{K} \vdash_P$ ，当且仅当 $\mathcal{K} \cup \{\neg\}$ 且 $_\mathcal{K}(\neg$ )=0 通过算法 3.1 得到一个完全的封闭的森林。

### 3.3 基于超协调表演算的推理

描述逻辑 *ALC* 中有两个基本推理问题：实例检查和概念包含。下面给出它们基于 $\vdash_P$ 关系的定义。

**定义** 3.5. 给定一个本体 *O*，一个概念 *C* 和一个个体 *a*，*a* 是 *C* 的关于 *O* 的实例当且仅当 $\mathcal{A} \vdash_P C(a)$。

**定义** 3.6. 给定一个本体 *O* 和概念 *C,D*， *C* 包含 *D* 当且仅当 $O \vdash_P C \sqsubseteq D$。

根据引理 1.1 和引理 1.2 可知，概念包含可以转换为实例检查问题，关于 TBox 推理问题可以转换为关于 ABox 推理问题。所以，本文只讨论在给定 ABox 中实例检查问题。

**定理** 3.5. 给定一个 ABox $\mathcal{A}$ ，一个概念 *C* 和一个个体 *a*，*a* 是 *C* 的关于 $\mathcal{A}$ 的实例当且仅当 $\mathcal{A} \cup \{\neg C(a)\}$ 且 $_\mathcal{A}(\neg C(a))=0$ 通过算法 3.1 得到一个完全的封闭的森林。

**证明**：由定义 3.4 可知，$\mathcal{A} \cup \{\neg C(a)\}$ 且 $_\mathcal{A}(\neg C(a))=0$ 通过算法 3.1 得到一个完全的封闭的森林当且仅当 $\mathcal{A} \vdash_P C(a)$；再由定义 3.5 可知 $\mathcal{A} \vdash_P C(a)$ 当且仅当 *a* 是 *C* 的关于 *O* 的实例。故，*a* 是 *C* 的关于 $\mathcal{A}$ 的实例当且仅当 $\mathcal{A} \cup \{\neg C(a)\}$ 且 $_\mathcal{A}(\neg C(a))=0$ 通过算法 3.1 得到一个完全的封闭的森林。
□

**注**：在下面例子中，为了方便记号，我们直接在一个概念左上角增加标记"0"和"1"来表示该公式的标记。

**例** 1. 设一个 ABox $\mathcal{A}$={*Penguin*(*Tweety*),¬*Fly*(*Tweety*), ¬*Swallow*(*Tweety*), ¬*Penguin*⊔*Fly*(*Tweety*)，¬*Swallow*⊔∀*HasFood*. ¬*Fish*(*Tweety*), ¬*Penguin* ⊔∃*HasFood*. *Fish* (*Tweety*), *HasFood*(*Tweety*, *Fingerling*)}。易知，$\mathcal{A}$ 是经典不协调的，下面通过算法 3.1 来对下面查询进行超协调推理。

（1） 查询 *Haswing*(*Tweety*)。

令 $\mathcal{A} \cup \{\neg Haswing(Tweety)\}$= {*Penguin*(*Tweety*)$^1$, ¬*Fly*(*Tweety*)$^1$, ¬*Penguin*⊔*Bird*(*Tweety*)$^1$, ¬*Bird*⊔*Fly*(*Tweety*)$^1$, ¬*Bird* ⊔∃*HasFood*. *Fish* (*Tweety*)$^1$, *HasFood*(*Tweety*, *Fingerling*)$^1$, ¬ *Haswing*(*Tweety*)$^0$}。通过算法 3.1 得到森林 $\_\mathcal{A}^1$={$L_{11}, L_{12}, L_{13}, L_{14}, L_{15}, L_{16}, L_{17}, L_{18}$} 如下：

$L_{11}$(*Tweety*)= {*Penguin*$^1$,¬*Fly*$^1$, ¬*Penguin*$^1$, ¬*Bird*$^1$, ¬ *Haswing*$^0$}。$L_{11}$ 包含实冲突 { *Penguin*$^1$, ¬*Penguin*$^1$}。

$L_{12}$ (*Tweety*)={*Penguin*$^1$,¬*Fly*$^1$, ¬*Penguin*$^1$, ¬*Bird*$^1$, ¬ *Haswing*$^0$}； $L_{12}(\iota)$={*Fish*$^1$}。

这里 $\iota$ 表示没有出现在 $\mathcal{A}$ 中的新个体(下同)。$L_{12}$ 包含实冲突 { *Penguin*$^1$, ¬*Penguin*$^1$}。

$L_{13}(Tweety)$= {$Penguin^1$,¬$Fly^1$, ¬$Penguin^1$, ¬$Fly^1$ , ¬$Bird^1$, ¬ $Haswing^0$}。

$L_{13}$ 包含实冲突{ $Penguin^1$, ¬$Penguin^1$}。

$L_{14}(Tweety)$= {$Penguin^1$,¬$Fly^1$, ¬$Penguin^1$, ¬$Fly^1$ ,¬ $Haswing^0$}；

$L_{14}(\iota)$={$Fish^1$}。$L_{14}$ 包含实冲突{ $Penguin^1$, ¬$Penguin^1$}

$L_{15}(Tweety)$= {$Penguin^1$,¬$Fly^1$, $Bird^1$, ¬$Bird^1$,¬ $Haswing^0$}。$L_{15}$ 包含实冲突{ ¬$Bird^1$, $Bird^1$}。

$L_{16}(Tweety)$= {$Penguin^1$,¬$Fly^1$, $Bird^1$, ¬$Bird^1$,¬$Haswing^0$}；$L_{16}(\iota)$={$Fish^1$}。

$L_{16}$ 包含实冲突{ $Bird^1$, ¬$Bird^1$}。

$L_{17}(Tweety)$= {$Penguin^1$,¬$Fly^1$, $Bird^1$,$Fly^1$,¬ $Bird^1$, ¬$Haswing^0$}。

$L_{17}$ 包含实冲突{ ¬$Fly^1$, $Fly^1$}和{ $Bird^1$, ¬ $Bird^1$}。

$L_{18}(Tweety)$= { $Penguin^1$,¬$Fly^1$, $Bird^1$, ¬$Bird^1$, ¬ $Haswing^0$}；$L_{18}(\iota)$={$Fish^1$}。

$L_{18}$ 包含实冲突{ ¬$Fly^1$, $Fly^1$}和{ $Bird^1$, ¬ $Bird^1$}。

因为分支 $L_{1i}$(i=1,…,8)都不包含第二类冲突，所以森林 $\rfloor_{\mathcal{A}}^1$ 不是封闭的。 由定理 3.5 可知，$\mathcal{A} \not\vdash_P$ $Haswing(Tweety)$，即 $Tweety$ 不是概念 $Haswing$ 的实例。既说明算法 3.1 是可以进行超协调推理又说明 $\mathcal{A}$ 中给出的关于 $Tweety$ 与 $Haswing$ 信息太少。

（2） 查询 $Fly(Tweety)$。

令 $\mathcal{A} \cup \{\neg Haswing(Tweety)\}$= {$Penguin(Tweety)^1$, ¬$Fly(Tweety)^1$, ¬$Penguin \sqcup Bird(Tweety)^1$, ¬$Bird \sqcup Fly(Tweety)^1$, ¬$Bird \sqcup \exists HasFood. Fish (Tweety)^1$, $HasFood(Tweety, Fingerling)^1$, ¬ $Fly(Tweety)^0$}。通过算法 3.1 得到森林 $\rfloor_{\mathcal{A}}^2$={$L_{21}$,$L_{22}$,$L_{23}$,$L_{24}$,$L_{25}$,$L_{26}$,$L_{27}$,$L_{28}$},如下：

$L_{21}(Tweety)$= {$Penguin^1$,¬$Fly^1$, ¬$Penguin^1$, ¬$Bird^1$, ¬ $Fly^0$}。$L_{21}$ 包含实冲突{ $Penguin^1$, ¬$Penguin^1$}。

$L_{22} (Tweety)$={$Penguin^1$,¬$Fly^1$, ¬$Penguin^1$, ¬$Bird^1$, ¬ $Fly^0$}； $L_{22}(\iota)$={$Fish^1$}。

$L_{22}$ 包含实冲突{ $Penguin^1$, ¬$Penguin^1$}。

$L_{23}(Tweety)$= {$Penguin^1$,¬$Fly^1$, ¬$Penguin^1$, ¬$Fly^1$ ,¬$Bird^1$,¬ $Fly^0$}。

$L_{23}$ 包含实冲突{ $Penguin^1$, ¬$Penguin^1$}。

$L_{24}(Tweety)$= {$Penguin^1$,¬$Fly^1$, ¬$Penguin^1$, ¬$Fly^1$ ,¬ $Fly^0$}；

$L_{24}(\iota)$={$Fish^1$}。$L_{24}$ 包含实冲突{ $Penguin^1$, ¬$Penguin^1$}。

$L_{25}(Tweety)$= {$Penguin^1$,¬$Fly^1$, $Bird^1$, ¬$Bird^1$,¬ $Fly^0$}。$L_{25}$ 包含实冲突{ ¬$Bird^1$, $Bird^1$}。

$L_{26}(Tweety)$= {$Penguin^1$,¬$Fly^1$, $Bird^1$, ¬$Bird^1$,¬ $Fly^0$}；$L_{26}(\iota)$={$Fish^1$}。

$L_{26}$ 包含实冲突{ $Bird^1$, ¬$Bird^1$}。

$L_{27}(Tweety)$= {$Penguin^1$,¬$Fly^1$, $Bird^1$,$Fly^1$,¬ $Bird^1$, ¬ $Fly^0$}。

$L_{27}$ 包含实冲突{ ¬$Fly^1$, $Fly^1$}和{ $Bird^1$, ¬ $Bird^1$}和强冲突{ $Fly^1$, ¬ $Fly$ }。

$L_{28}(Tweety)$= { $Penguin^1$,¬$Fly^1$, $Bird^1$, ¬$Bird^1$, ¬ $Fly^0$}；$L_{28}(\iota)$={$Fish^1$}。

$L_{28}$ 包含实冲突{ $Bird^1$, ¬ $Bird^1$}。

因为分支 $L_{27}$ 包含第二类冲突，所以森林 $\rfloor_{\mathcal{A}}^2$ 是封闭的。 由定理 3.5 可知，$\mathcal{A} \vdash_P Fly(Tweety)$，即 $Tweety$ 是概念 $Fly$ 的实例。当再查询¬$Fly(Tweety)$，同理得到 $L_{27}$ 都包含第二类冲突，所以森林 $\rfloor_{\mathcal{A}}^2$ 是封闭的。由定理 3.5 可知，$\mathcal{A} \vdash_P$ ¬$Fly(Tweety)$，即 $Tweety$ 是概念 ¬$Fly$ 的实例。既说明算法 3.1 对不协调的信息以矛盾形式给出。

（3） 查询$\exists HasFood. Fish (Tweety)$

令 $\mathcal{A} \cup \{\neg Haswing(Tweety)\}$= { $Penguin(Tweety)^1$, ¬$Fly(Tweety)^1$, ¬$Penguin \sqcup Bird(Tweety)^1$, ¬$Bird \sqcup Fly(Tweety)^1$, ¬$Bird \sqcup \exists HasFood. Fish (Tweety)^1$, $HasFood(Tweety, Fingerling)^1$, ¬ $\exists HasFood. Fish (Tweety)^0$}。通过算法 3.1 得到森林 $\rfloor_{\mathcal{A}}^3$={$L_{31}$,$L_{32}$,$L_{33}$,$L_{34}$,$L_{35}$,$L_{36}$,$L_{37}$,$L_{38}$},如下：

$L_{31}(Tweety)$= {$Penguin^1$,¬$Fly^1$, ¬$Penguin^1$, ¬$Bird^1$}；$L_{31}(Fingerling)$={ ¬$Fish^0$}。

$L_{31}$ 包含实冲突{ $Penguin^1$, ¬$Penguin^1$}。

$L_{32} (Tweety)$={$Penguin^1$,¬$Fly^1$, ¬$Penguin^1$, ¬$Bird^1$}; $L_{31}(Fingerling)$={ ¬$Fish^0$}； $L_{32}(\iota)$={$Fish^1$, ¬$Fish^0$}。

$L_{32}$ 包含实冲突 $\{Penguin^1, \neg Penguin^1\}$ 和强冲突 $\{Fish^1, \neg Fish^0\}$。

$L_{33}(Tweety) = \{Penguin^1, \neg Fly^1, \neg Penguin^1, \neg Fly^1, \neg Bird^1\}$；$L_{33}(Fingerling) = \{\neg Fish^0\}$

$L_{33}$ 包含实冲突 $\{Penguin^1, \neg Penguin^1\}$。

$L_{34}(Tweety) = \{Penguin^1, \neg Fly^1, \neg Penguin^1, \neg Fly^1\}$；$L_{34}(Fingerling) = \{\neg Fish^0\}$；$L_{34}(\iota) = \{Fish^1, \neg Fish^0\}$。

$L_{34}$ 包含实冲突 $\{Penguin^1, \neg Penguin^1\}$ 和强冲突 $\{Fish^1, \neg Fish^0\}$。

$L_{35}(Tweety) = \{Penguin^1, \neg Fly^1, Bird^1, \neg Bird^1\}$；$L_{34}(Fingerling) = \{\neg Fish^0\}$。$L_{35}$ 包含实冲突 $\{\neg Bird^1, Bird^1\}$。

$L_{36}(Tweety) = \{Penguin^1, \neg Fly^1, Bird^1, \neg Bird^1\}$；$L_{36}(Fingerling) = \{\neg Fish^0\}$；$L_{36}(\iota) = \{Fish^1, \neg Fish^0\}$。

$L_{36}$ 包含实冲突 $\{Bird^1, \neg Bird^1\}$ 和强冲突 $\{Fish^1, \neg Fish^0\}$。

$L_{37}(Tweety) = \{Penguin^1, \neg Fly^1, Bird^1, Fly^1, \neg Bird^1\}$；$L_{37}(Fingerling) = \{\neg Fish^0\}$。

$L_{37}$ 包含实冲突 $\{\neg Fly^1, Fly^1\}$ 和 $\{Bird^1, \neg Bird^1\}$ 和强冲突 $\{Fly^1, \neg Fly\}$。

$L_{38}(Tweety) = \{Penguin^1, \neg Fly^1, Bird^1, \neg Bird^1\}$；$L_{38}(Fingerling) = \{\neg Fish^0\}$；$L_{38}(\iota) = \{Fish^1, \neg Fish^0\}$。

$L_{38}$ 包含实冲突 $\{Bird^1, \neg Bird^1\}$ 和强冲突 $\{Fish^1, \neg Fish^0\}$。

因为分支 $L_{3i}(i=2,4,6,8)$ 都包含第二类冲突，所以森林$\_\mathcal{A}^3$ 是封闭的。由定理 3.5 可知，$\mathcal{A} \vdash_P \exists HasFood.Fish(Tweety)$，即 Tweety 是概念 $\exists HasFood.Fish$ 的实例。

例 1 表明，这种基于符号变换所定义的封闭条件，使矛盾的信息在推理的过程中，没有被传递，很好地被局部化，从而有效地避免了平凡扩展以达到处理不协调信息的目的。

## 4 超协调表演算的性质

我们可以看出这种基于符号变换的超协调算法是可终止的。

**定理** 4.1. 算法 3.1 是可终止的。

**证明**：设 $m=|clos(\mathcal{A})|$，$n=|R_\mathcal{A}|$。要证明算法 3.1 是可终止的，只需证明应用扩展规则满足下列性质：

（1）所有扩展规则的使用不会从森林$\_\mathcal{A}$中删除任何结点。由扩展规则定义可以直接得出。

（2）森林$\_$的深度是有界的。

对森林$\_\mathcal{A}$中每个结点 x 定义 $l(x)$ 为 $L(x)$ 中概念所包含的 ∃ 和 ∀ 的个数的最大值。显然对所有的 K 中出现的个体 $x_0$，有 $l(x_0) < |K|$。同时对于任意的边 $<x,y>$，$l(x) > l(y)$。于是对任意一条路径 $<x_1,...,x_k>$，有 $l(x_1) > ... > l(x_k)$。于是森林$\_\mathcal{A}$的深度可以被 $|K|$ 界定。

（3）森林$\_\mathcal{A}$的广度是有界的。算法 3.1 仅使用 ∃-规则从 $clos(\mathcal{A})$ 中的形如 ∃R.C 的概念产生一个新结点。因为 $clos(\mathcal{A})$ 包含至多 m 个形如 ∃R.C 的概念，所以森林$\_\mathcal{A}$的出度的最大界是 mn，即，森林$\_$的广度是有界的，而且可以被 $|K|$ 界定。

通过（1），（2）和（3）得出，算法 3.1 是可终止的。□

我们的超协调表演算在处理协调的本体的能力与经典表演算是一样的。

**定理** 4.2. 给定一个协调的 ABox $\mathcal{A}$ 和一个断言 $C(a)$，$\mathcal{A} \vdash_P C(a)$ 当且仅当 $\mathcal{A} \vDash C(a)$。

**证明**：（充分性）如果 $\mathcal{A} \vdash_P C(a)$ 那么由定理 3.5 可知，令 $S=\mathcal{A} \cup \{\neg C(a)\}$ 通过算法 3.1 生成的森林$\_s$是一个封闭的，即根据定义 3.3 中的（6）和（7）可知，任意的树都是都包含冲突而且存在一个树包含第二类冲突。因为 $\mathcal{A}$ 是一个协调的 ABox 和森林$\_s$每一棵树都包含冲突，所以 $\mathcal{A} \vDash C(a)$，由定理（参加 DL handbook）。

（必要性）令 $S=\mathcal{A} \cup \{\neg C(a)\}$，根据通过算法 3.1 生成的一个完全的森林$\_s$。只需要证明在 $\mathcal{A}$ 是一个协调的 ABox 条件下，森林$\_s$是封闭的。首先如果 $\mathcal{A} \vDash C(a)$，因为 $\mathcal{A} \cup \{\neg C(a)\}$ 是不可满足的，所以森林$\_s$的每个树都包含冲突。假设森林$\_s$所有树不包含第二类冲突，即所有的冲突公式都来源于 $\mathcal{A}$ 中而与 $\neg C(a)$ 无关。于是得出 $\mathcal{A}$ 是不协调的，矛盾。所以假设不成立，森林$\_s$存在一个树包含第二类冲突。故，森林$\_s$是封闭的，即 $\mathcal{A} \vdash_P C(a)$。□

这样的符号变换不改变推理问题的复杂度，即，

**定理** 4.3. 给定一个 *ALC* 本体 *O*，一个概念 *C* 和一个个体 *a*，判定 $\mathcal{A} \vdash_P C(a)$ 是否成立的复杂度是 PSPACE-完全的。

**证明**：

（1）存在 PSPACE 的算法

根据定理 4.1 的证明，我们已经看到 $\lrcorner_\mathcal{A}$ 的大小是可以被 |K| 的多项式界定的，于是算法 3.1 是一个 PSPACE 的算法。

（2）任何一个 PSPACE 问题都可以归约到它。

任给一个 ALC 的概念 C，判定 C 的不可满足性问题是 PSPACE 完全的[16]。我们下面证明判定 C 的不可满足性可以归约到超协调表演算。

令 A 是一个不在 C 中出现的原子概念。C 是不可满足的当且仅当 $\{C \sqcup A(a)\} \vDash A(a)$。$\{C \sqcup A(a)\}$ 是协调的，因为可以构造一个解释 $I=(\Delta^I, \cdot^I)$，其中 $\Delta^I = \{a\}$，$a^I = a$，$A^I = \{a\}$。根据定理 4.2，$\{C \sqcup A(a)\} \vDash A(a)$ 当且仅当 $\{C \sqcup A(a)\} \vdash_P A(a)$。这样我们就完成了归约。□

当本体不协调的时候，经典表演算会推出任意的结论，而我们的表演算具有超协调的能力，即

**定理** 4.4. $\vdash_P$ 是超协调的。

**证明**：对任意的概念 C 和个体 a，令 $\mathcal{A}=\{C(a), \neg C(a)\}$。只需要证明 $\mathcal{A} \nvdash_P D(a)$，对任意不同于 C 概念 D。令 $\mathcal{A} \cup \{\neg D(a)\} = \{C(a), \neg C(a), \neg D(a)\}$ 且 $\lrcorner_\mathcal{A}(C(a)) = \lrcorner_\mathcal{A}(\neg C(a)) = 1$，$\lrcorner_\mathcal{A}(\neg D(a))=0$。设 $\lrcorner_\mathcal{A}$ 通过算法 3.1 而得到的森林，尽管 $\lrcorner_\mathcal{A}$ 包含实冲突 $\{A(a), \neg A(a)\}$ 且 $\lrcorner_\emptyset(A(a)) = \lrcorner_\emptyset(\neg A(a)) = 1$，但是 $\lrcorner_\mathcal{A}$ 不包含第二类冲突，所以 $\lrcorner_\mathcal{A}$ 是封闭的。由定义 3.4 可知，$\mathcal{A} \nvdash_P D(a)$。故，$\vdash_P$ 是超协调的。□

对于重言式，在我们的超协调表演算下，也是成立的。

**定理** 4.5. $\emptyset \vdash_P \top(a)$。

**证明**：考虑 $\emptyset \cup \{\neg \top(a)\}$ 且 $\lrcorner_\emptyset(\neg \top(a))=0$。设 $\lrcorner_\emptyset$ 通过算法 3.1 而得到的森林，$\lrcorner_\emptyset$ 包含强冲突 $\{A(a), \neg A(a)\}$ 且 $\lrcorner_\emptyset(A(a)) = \lrcorner_\emptyset(\neg A(a))=0$，因为 $\neg \top \equiv \bot$ 且 $\bot \equiv A \sqcap \neg A$，这里 A 是一个新的原子概念。

由定理 3.3 可知，$\emptyset \vdash_P \top(a)$。□

# 5 相关工作

本文提出基于符号变换的描述逻辑表演算是一种超协调处理方法。接下来，我们将比较基于符号变换的超协调表演算与目前在描述逻辑中超协调处理的几个主要工作(文献[8,9,10,11,15])做比较。

与四值描述逻辑[8,15]的推理系统相比，基于符号变换的超协调表演算克服了四值推理系统不满足三条基本推理规则的缺点，从而使得推理能力大大加强。直观上，基于符号变换的超协调表演算并不像四值描述逻辑推理系统把矛盾的信息以矛盾的形式隔离起来，而是让不协调信息中成真的部分参与到推理中来。从这种意义上讲，基于符号变换的超协调表演算大大提高了信息的利用价值。

与准经典描述逻辑[10,11]的推理系统相比，基于符号变换的超协调表演算克服了准经典描述逻辑的相关性的缺点（即不满足排中律），从而有效地推理出所有重言式(真理)。另外，准经典描述逻辑的推理系统是通过限制推理规则的使用而避免了平凡扩展。而基于符号变换的超协调表演算不限制推理规则的使用次序，推理出的结论更具可信。

与概率描述逻辑[9]等量化形式来实现推理相比，基于符号变换的超协调表演算是通过定性方式来实现推理。量化方式会带来非确定性，例如，推理的结论非确定性。与概率描述逻辑修改了描述逻辑的语法结构相比，而基于符号变换的超协调表演算保持描述逻辑的经典语法。

总之，与其它的在描述逻辑中超协调处理的方法相比，基于符号变换的超协调推理方法在不改变描述逻辑原始语法，而是在经典推理系统之外增加了一层矛盾冲突分析系统，从而实现超协调处理。我们的

超协调方法比四值逻辑和准经典逻辑具有更强的推理能力而且保持经典描述逻辑中许多优良的性质。

# 6 结束语

语义万维网一开始就肩负着改造现有万维网的重任，它正在逐渐改变和影响我们现有的万维网。因为语义万维网的本体可能是分布式的，可能是多作者的，可能是由不同的数据源得来，所以在语义万维网环境下，真实的应用数据一般来讲是很容易包含矛盾信息[14]。然而，作为语义万维网的逻辑基础—描述逻辑不能处理不协调，所以在描述逻辑本体中处理不协调问题越来越引起计算机领域的重视。本文在启发于悖论逻辑的表演算[14,17]基础上，为矛盾冲突分成了三个种类：实冲突、强冲突和内冲突。通过这样的三种冲突来定义表演算的封闭条件，从而实现超协调推理的任务。这样，我们可以通过设定封闭条件就能达到实现不同推理的目的。在本文中，我们把强冲突和内冲突设定为封闭条件，使得基于符号变换的描述逻辑表演算满足三条基本推理规则(分离规则(MP)，拒取推理(MT)和析取三段论(DS))和排中律，从而使得基于符号变换的描述逻辑表演算具有近似经典逻辑系统的推理能力。我们提出的基于冲突分类的处理不协调方法是用来处理描述逻辑中的不协调问题一种新的尝试。当前，在描述逻辑中研究不协调处理的推理机已是语义网研究领域一个重要的前沿问题。下一步研究的主要目标是基于这个超协调表演算而建立一个描述逻辑的超协调推理机。

**References**:

[1] Tim Berners-Lee. Semantic Web. September 1998, http://www.w3.org/DesignIssues/Semantic.html
[2] Tim Berners-Lee,James Hendler, Ora Lassila. The Semantic Web. Scientific American, 284(5):35~43,2001
[3] Nigel Shadbolt, Tim Berners-Lee, Wendy Hall. The Semantic Web revisited. *IEEE Intelligent Systems*, 21(3):96-101, 2006
[4] Tim Berners-Lee, Dan Connolly, Lalana Kagal, Yosi Scharf, Jim Hendler. N3logic. A logical framework for the World Wide Web. *CoRR*,abs/0711.1533, 2007
[5] Lalana Kagal,Tim Berners-Lee,Dan Connolly, Daniel J.Weitzner. Using Semantic Web technologies for policy management on the web. In: *Proceedings of the 21th National conference on Artificial Intelligence and the 8th Innovative Applications of Artificial Intelligence Conference(AAAI/IAAI'06), Boston, Massachusetts, USA, 2006.* AAAI Press(2006).
[6] Schlobach,S., Cornet,R. Non-standard reasoning services for the debugging of description logic terminologies. In:*Proceedings of the 8th International Joint Conference on Artificial Intelligence (IJCAI'03),Acapulco,Mexico,2003*. Morgan Kaufmann, 355~362,2003
[7] Huang,Z., van Harmelen,F., ten Teije, A. Reasoning with inconsistent ontogies. In:*Proceedings of the 9th International Joint Conference on Artificial Intelligence (IJCAI'05),Edinburgh, Scotland,UK, 2005*. Professional Book Center, 454~459,2005
[8] Ma,Y., Hitzler,P., Lin,Z. Paraconsistent resolution for four-valued description logics. In:*Proceedings of the 4th European Semantic Web Conference (ESWC'07).Innsbruck, Austria*, LNCS 250. Springer, 399~413,2007
[9] Qi,G., Pan,J.Z., Ji,Q. Extending description logics with uncertainty reasoning in possibilistic logic. In:*Proceedings of the 9th European Conference on Symbolic and Quantitative Approaches to Reasoning with Uncertainty (ECSQARU'07), Hammamet, Tunisia*. LNCS 4724, Springer, 828-839,2007
[10] Zhang,X., Lin, Z. Paraconsistent reasoning with quasi-classical semantics in ALC. In:*Proceedings of the 2nd International Conference of Web Reasoning and Rule Systems (RR'08), Karlsruhe, Germany*. LNCS 5341, Springer, 222~229,2008
[11] Zhang,X., Guo, X., Lin, Z. A Tableau Algorithm for Handling Inconsistency in OWL. In:*Proceedings of the 6nd European Semantic Web Conference (ESWC'09), 31 May - 4 June 2009, Heraklion, Greece*. LNCS 5554, Springer, 399~413,2009
[12] Baader,F., Calvanese,D., McGuiness,D.,Nardi,D.,Patel-Schneider,P., eds.  The description Logic in Handbook: Theory, Implementation, and Applications. *Cambridge University Press*, 2003.
[13] Horrocks,I., Sattler,U., Tobies,S. Reasoning with individuals for the description logic SHIQ. *CoRR* cs.LO/005017,2000

[14] Zuoquan Lin, Tableau Systems for Paraconsistency and Minimal Inconsistency, *Journal of Computer Science and Technology*，Vol.13,No.2,174~188，1998

[15] Ma Yue. Dealing with inconsistencies in the Semantic Web. [Ph.D.Thesis]. *Beijing: Peking University*, 2008(in Chinese with English abstract)

[16] Manfred Schmidt-Schauß and Gert Smolka. Attributive concept descriptions with complements. *Artificial Intelligence*, 48(1):1-26, 1991.

附中文参考文献:

[15] 马跃. 语义万维网中的不协调知识处理[博士学位论文]. 北京:北京大学. 2008

[17] 林作铨. 李未. 悖论逻辑的表演算.软件学报, 1996, 7 (06): 345-353